\documentclass[a4paper,conference]{IEEEtran}
\ifCLASSINFOpdf
\else
\fi

\usepackage{amsmath}
\usepackage{amsfonts}
\usepackage{graphicx}
\usepackage{subcaption}
\usepackage{url}
\usepackage{multicol}
\usepackage{makecell}
\usepackage{xspace}
\usepackage[dvipsnames]{xcolor}
\usepackage{stfloats}
\begin{document}
%
\title{MontageGAN: Generation and Assembly of Multiple Components by GANs}


\author{\IEEEauthorblockN{Chean Fei Shee}
\IEEEauthorblockA{Kyushu University\\
Fukuoka, Japan\\
Email: shee.c.fei@human.ait.kyushu-u.ac.jp}
\and
\IEEEauthorblockN{Seiichi Uchida}
\IEEEauthorblockA{Kyushu University\\
Fukuoka, Japan\\
Email: uchida@ait.kyushu-u.ac.jp}}


%


\maketitle

\newcommand\red[1]{\color{black}#1\xspace\color{black}}
\newcommand\green[1]{\color{black}#1\xspace\color{black}}
\newcommand\todo[1]{\color{black}#1\xspace\color{black}}
\addtolength\textfloatsep{-12.5pt}

\begin{abstract}
A multi-layer image is more valuable than a single-layer image from a graphic designer's perspective.
However, most of the proposed image generation methods so far focus on single-layer images.
In this paper, we propose MontageGAN, which is a Generative Adversarial Networks (GAN) framework for generating multi-layer images.
Our method utilized a two-step approach consisting of local GANs and global GAN.
Each local GAN learns to generate a specific image layer, and the global GAN learns the placement of \green{each generated image layer}.
Through our experiments, we show the ability of our method to generate multi-layer images and estimate the placement of the generated image layers.
\end{abstract}


%
\IEEEpeerreviewmaketitle

\section{Introduction}

\textit{Layering} is one of the most critical concepts in graphic design that has become a must-have feature in popular raster graphic editors such as \textit{Adobe Photoshop} \cite{photoshop}, \textit{GIMP} \cite{gimp}, and \textit{Krita} \cite{krita}.
In graphic design, layering is the stacking of different images, graphics, text, etc., on top of one another.
A multi-layer image such as \textit{PSD} file (Photoshop Document) and \textit{XCF} file (eXperimental Computing Facility, used by GIMP) is more valuable than a simple single-layer image such as \textit{JPEG} file and \textit{PNG} file, in graphic designers' perspective.
Layering makes it easier for designers to design and edit images because they can apply changes to only one particular layer without affecting the other layers.
Besides that, layering allows designers to show or hide specific layers to create a slightly altered design version.

Designing an image can be time-consuming and takes a lot of effort.
Therefore, there is a need for tools that can automatically and quickly generate images with novel designs.
Recently, unsupervised learning with \textit{Generative Adversarial Networks} (GAN) \cite{goodfellow2014generative} to learn the deep representation of the training data has gained popularity, especially in image generation.
State-of-the-art methods, for instance, \textit{StyleGAN} \cite{karras2019style} (and its improved models, \textit{StyleGAN2} \cite{karras2020analyzing} and \textit{StyleGAN2-ada} \cite{karras2020training}) and \textit{Projected GANs} \cite{sauer2021projected}, yield significant success in generating not only photo-realistic images such as human faces, buildings, and animals, but also illustration images such as paintings and digital artworks.

We focus on multi-layer image generation using GAN and explore its potential in this work.
Despite GAN's notable success in image generation, most of the proposed methods focus on single-layer image generation, limiting usability and flexibility in graphic design.
Unlike traditional image generation, which outputs single-layer images, we generate multiple image layers with alpha channels to handle transparency.
These image layers are then alpha-blended finally to form the target overlaid image. 

As the target images, we consider the front face of animation (also known as \textit{Anime}) characters.
Specifically, our objective is to generate every face part like eyes, nose, mouth, and hair in separate image layers.
As mentioned earlier, generating Anime characters in multi-layer images has several advantages.
For example, multi-layer images make it relatively easier to animate the characters afterward than traditional single-layer images.
It is also possible to manipulate the character design by swapping between layers generated using different latent variables, achieving layer-wise manipulation.

\begin{figure}[t]
    \centering
    \includegraphics[width=0.8\linewidth]{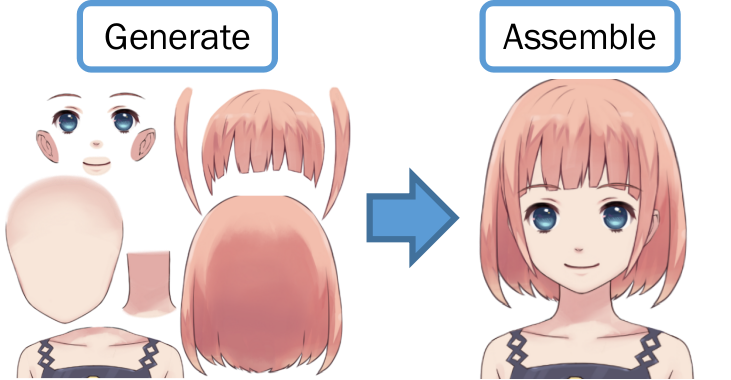}
    \caption{The concept of multi-layer image generation. The image of this character is retrieved from Live2D's sample \cite{live2dsample}.}
    \label{fig:concept}
\end{figure}

We propose a GANs framework, called \textit{MontageGAN}, that generates multi-layer Anime face images.
Fig.\ref{fig:concept} shows our concept of multi-layer image generation.
MontageGAN utilizes a two-step approach consisting of \textit{local GANs} and a \textit{global GAN}.
Each local GAN based on StyleGAN2-ada learns to generate a specific face part; for example, the local GAN of eyes generates only the image layers of the eyes.
On the other hand, the global GAN learns to assemble all image layers generated by local GANs into a reasonably-looking face.
Note that we aim not only to place the generated parts into the correct position but also to generate layers that match the styles of each other (\textit{global styles}).
In addition to the benefits of generating multi-layer images, we believe that splitting image generation at the layer level will make learning simple, as each local GAN focuses only on its layer.

\pagebreak
The main contributions of this paper are summarized as follows:
\begin{itemize}
    \item As far as the authors know, this is the first image generation model specializing in multi-layer image generation.
    With both local GANs and global GAN, our method mimics the process of designing multi-layer images.
    \item We proposed a GANs framework and applied it to Anime face image generation by training the model with a multi-layer image dataset.
\end{itemize}
Our implementation is available at \url{https://github.com/uchidalab/docker-montage-gan}.
\section{Related Work}

\subsection{Generative Adversarial Networks}
As its name suggests, GAN utilizes two adversarial neural networks, usually called the \textit{generator} and the \textit{discriminator} \cite{goodfellow2014generative}.
The discriminator learns to determine if the input data was generated or sampled from the training data distribution.
On the other hand, the generator learns to generate a fake that fools the discriminator into believing it is a real one.

Much effort has been made to stabilize GAN training and improve the quality of the generated images \cite{arjovsky2017wasserstein, gulrajani2017improved, mao2017least, mescheder2018training}.
A recent state-of-the-art method, \textit{StyleGAN} \cite{karras2019style} (and the improved model, \textit{StyleGAN2} \cite{karras2020analyzing}), proposed a redesign of generator architecture motivated by style transfer literature, which has notably improved the resulting quality and better disentanglement.
The StyleGAN generator's architecture differs from the traditional GANs; it consists of 2 modules: \textit{mapping} and \textit{synthesis networks}.
The mapping network is a \textit{Multilayer Perceptrons} (MLP) that learns a mapping $f \colon \mathcal{Z} \mapsto \mathcal{W}$ from latent variables $z \in \mathcal{Z}$ to intermediate latent variables $w \in \mathcal{W}$.
Conventionally, the latent space $\mathcal{Z}$ had to follow the probability density of the training data, leading to some unavoidable entanglement.
StyleGAN suggested that the intermediate latent space $\mathcal{W}$ has no such limitation, thus improving the disentanglement.
The synthesis network is a \textit{Convolutional Neural Network} (CNN) that generates images from a learned constant input.
The intermediate latent variables control the generator
through \textit{Adaptive Instance Normalization} (AdaIN) \cite{huang2017arbitrary} at each convolution layer of the synthesis network.
Also, Gaussian noise is added after each convolution layer to generate stochastic details.

Yet, training a high-quality GAN requires a sufficiently large dataset, usually around \(10^5\) to \(10^6\) images.
Often the problem of GAN with a small dataset is that the discriminator overfits the training samples.
The discriminator's feedback to the generator becomes meaningless, causing the training begins to diverge.
\textit{StyleGAN2-ada} \cite{karras2020training} proposed a wide range of \textit{non-leaking augmentations} to prevent the discriminator from overfitting.
The StyleGAN2-ada discriminator is evaluated using only augmented images and does this also when training the generator.
StyleGAN2-ada shows comparable results on several datasets, despite using a few thousand training images only.
In this work, we create the local GANs based on StyleGAN2-ada.

\subsection{Spatial Transformer Networks}
\textit{Spatial Transformer Networks} (STN) \cite{jaderberg2015spatial} is a learnable module that allows the spatial manipulation of data within the network.
STN can be injected into standard neural network architectures, allowing them to spatially transform the feature maps conditional to the feature maps itself, without extra supervision or modification to the current optimization process.
STN was initially used to increase the robustness of the image classification model \cite{he2019optimized, he2018lidar, wang2018geometry, park2018classification}; the image classification model using STN learns invariance to translation, scale, rotation, and more generic warping.

STN consists of 3 parts: \textit{localization network}, \textit{grid generator}, and \textit{sampler}.
The localization network is typically a CNN that predicts the transformation parameters $\theta$ based on the input image (or feature map).
After that, the grid generator creates a sampling grid $\mathcal{T}_\theta(G_{grid})$ (a set of points where the input image needs to be sampled to produce the transformed output image) based on the predicted transformation parameters.
Finally, the input image and the sampling grid are inputted to the sampler, creating the output image sampled from the input at the grid points.
Equation \eqref{eq:stn} show the coordinate mapping relationship of a 2D affine transformation $\mathbf{A}_\theta$, where $(x_s,y_s)$ and $(x_t,y_t)$ are the source coordinates and target coordinates, respectively.
In this work, we extend the STN's architecture to support multi-layer images and apply it to estimate the placement of parts in layer images.

\begin{equation}
\label{eq:stn}
    \binom{x_s}{y_s}=\mathcal{T}_\theta(G_{grid})=\mathbf{A}_\theta\begin{pmatrix}
x_t\\ 
y_t\\ 
1
\end{pmatrix}=\begin{bmatrix}
\theta_{11} & \theta_{12} & \theta_{13} \\
\theta_{21} & \theta_{22} & \theta_{23}
\end{bmatrix}\begin{pmatrix}
x_t\\ 
y_t\\ 
1
\end{pmatrix}
\end{equation}

\subsection{Differentiable Rendering}
\textit{Differentiable rendering} (DR) is a set of techniques that allows the gradients of 3D objects to be calculated and propagated through images \cite{kato2020differentiable}.
By differentiating the rendering, neural networks can optimize 3D entities (materials, lights, cameras, geometry, etc.) while operating on 2D projections (RGB image, depth image, silhouette image, etc.).
\textit{Neural rendering} is one DR technique that learns from data instead of handcrafting the rendering differentiation \cite{tewari2020state}.
There are various applications of neural rendering such as semantic photo manipulation, novel view synthesis, facial and
body reenactment, etc.
In this work, we consider multi-layer images as a type of 3D data, as the image layers stack along the z-axis to form a voxel.
Based on the neural rendering method, we first render the multi-layer image into a single-layer image before being input to the global discriminator.
It allows MontageGAN to generate multi-layer images, but as with typical GANs, the global discriminator can still make predictions based on the rendered single-layer images.

\begin{figure*}[t]
  \centering
  \includegraphics[width=\textwidth]{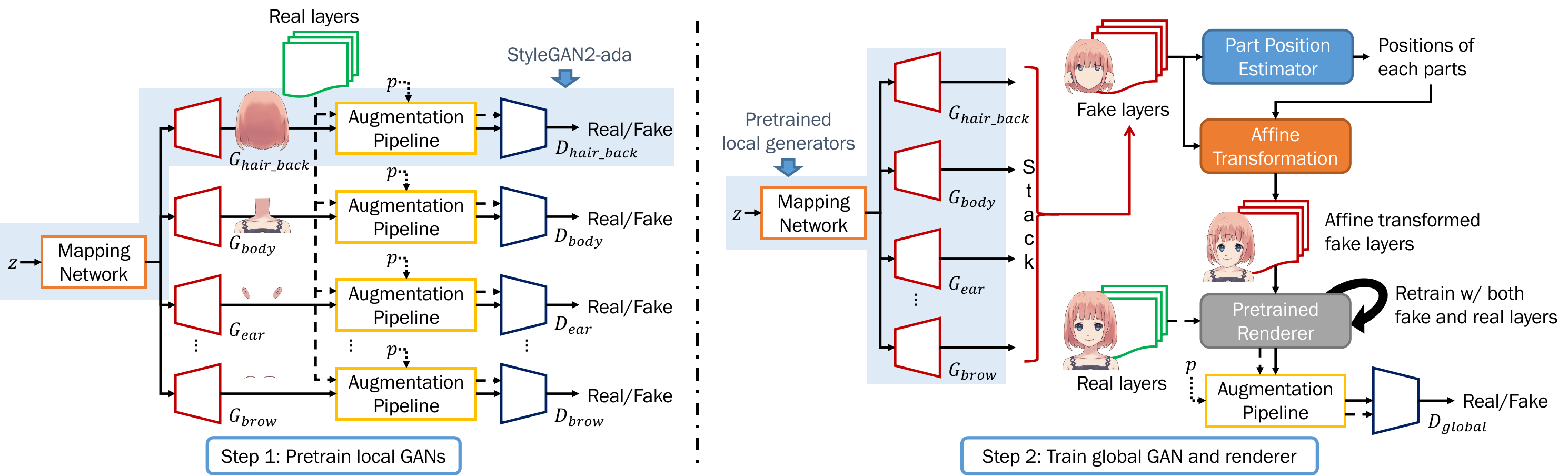}
  \caption{The overall architecture of MontageGAN. The left side shows step 1 (pretrain local GANs). The shaded area represents a single StyleGAN2-ada model. The right side shows step 2 (train global GAN and renderer). The shaded area represents the local generators that was pretrained in step 1.}
  \label{fig:merge-h}
\end{figure*}

\subsection{Anime Character Generations}
There are several attempts to use generative models in Anime character generation, albeit for not being a multi-layer image generation task that MontageGAN is tackling.
For example, a website called ``\textit{This Waifu Does Not Exist}" \cite{twdne} provides a demo that utilizes StyleGAN2 in Anime character generation.
Besides that, websites that offer interactive Anime character generation that users can finetune the parameters are also available, for instance, \textit{Artbreeder} \cite{artbreeder} and \textit{WaifuLabs} \cite{waifulabs}.
A style-guided generative model can generate an image that matches the desired style.
For example, \textit{AniGANs} \cite{li2021anigan} is a style-guided GAN that translates a human face into an Anime character based on the styles of a reference Anime character.
In addition, an article titled ``\textit{Talking Head Anime from a Single Image}" \cite{talkingheadanime} proposes a generative model that takes an Anime character's face and the desired pose as input and generates another image of the same character in the given pose.
\section{MontageGAN}

\begin{figure}[t]
    \centering
    \includegraphics[width=0.8\linewidth]{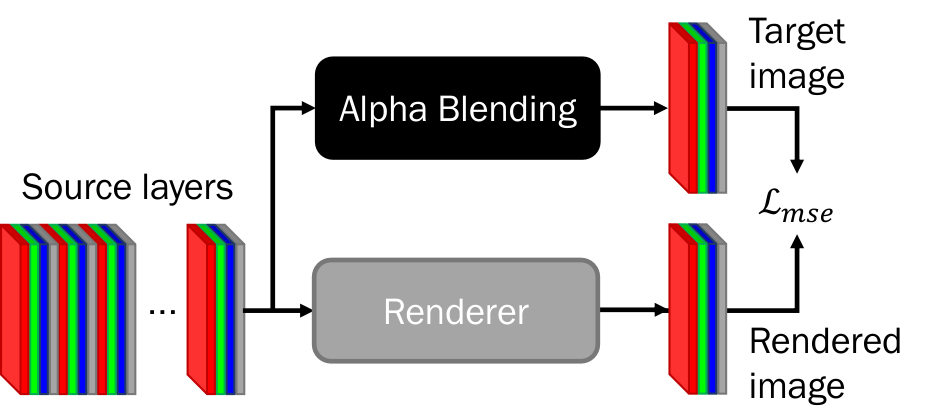}
    \caption{The training method of the renderer.}
    \label{fig:renderer}
\end{figure}

In this work, we generate multi-layer Anime character images with a two-step approach.
First, the local GANs generate the image layers that correspond to face features.
Then, the global GAN assembles the generated image layers into a finished image.
Fig. \ref{fig:merge-h} shows the overall architecture of MontageGAN.
$z \sim \mathcal{N}(0,1)$ denotes a random latent variable fed into the \red{mapping network~\cite{karras2020training}} to output an intermediate latent variable.
$G_i$ and $D_i$ denote the generator and discriminator, respectively, while the subscript $i$ indicates the image layer to which they belong.
Likewise, $D_{global}$ denotes the global discriminator. $p$ denotes the parameter of StyleGAN2-ada's \red{augmentation pipeline~\cite{karras2020training}} to control the augmentation probability.

\subsection{Local GANs}
We create the local GANs based on StyleGAN2-ada, with several modifications on top of the official source code\footnote{Accessible at \url{https://github.com/NVlabs/stylegan2-ada-pytorch}}.
\begin{enumerate}
    \item Training with RGBA channel images is possible. (As of writing, the official StyleGAN2-ada only supports grayscale and RGB images.)
    \item Training with rectangular-shaped images is possible as long as the dimension of both edges has a common divisor that can be represented as $2^n$, where $n\geq2$ and $n\in\mathbb{N}$. (As of writing, the official StyleGAN2-ada only supports square images.)
\end{enumerate}
We use a shared mapping network between the synthesis networks of the local generators.
The mapping network is a 8-layer MLP that outputs intermediate latent variables containing certain styles' parameters.
As local generators should generate images correlated with other layers, we believe there is no reason to use a separated mapping network.
We deploy a dedicated augmentation pipeline for \green{each local GAN}.
The \red{augmentation pipelines~\cite{karras2020training}} in local GANs augment the images before passing them to the local discriminators.
The applied augmentation prevents the local discriminators from overfitting, controlled by an augmentation probability.
\green{Each local GAN} can have a different target resolution to save computational resources; for instance, the local GANs that generate smaller face parts (mouth, nose, etc.) can target smaller resolutions.

\subsection{Renderer}
We create the differentiable renderer based on the neural rendering method.
Specifically, the renderer utilizes a 5-layer \textit{Fully Convolutional Network} (FCN) as the neural rendering network.
In this work, the renderer learns to blend the multi-layer images $I_{B,L \times C,H,W}$ into single-layer images $I_{B,C,H,W}$ by approximating the alpha blending operation. 
Where $B$, $L$, $C$, $H$, $W$ denote the batch size, number of image layers, number of image channels, image height, and image width, respectively.
Fig. \ref{fig:renderer} shows the training method of the renderer.
As the source layers, we use the real layers with random translation transformation applied during the pretraining.
The source layers are then blended with both alpha blending operation and the renderer, producing the target images and rendered images.
Finally, the pixelwise \textit{MSE loss} between the target and rendered images is calculated and back-propagated to the renderer for optimization.

In theory, the global discriminator still can make predictions directly based on multi-layer images without a renderer, but it tends to overfit quickly.
We believe that the global discriminator can easily distinguish between real and fake layers because there is too much information in the multi-layer images.
For example, a sample with a poorly generated layer that does not necessarily affect the overall appearance that much could be enough for the overfitted global discriminator to predict it as fake.
Hence, the render plays the role of limiting such information by only feeding the rendered images to the global discriminator.

\subsection{Global GAN}
The local generators and the \textit{part position estimator} together play the role of global generator in global GAN.
The local generators are given sufficient pretraining in step 1.
The generated image layers $\left \{I_{B,C,H_1,W_1}, \dots, I_{B,C,H_L,W_L}\right \}$ are zero-padded to the largest target resolution $(H,W)=(\max_{i\in{1,\dots,L}}H_i, \max_{i\in{1,\dots,L}}W_i)$ and stacked along the channel axis as fake layers $I_{B,L \times C,H,W}$.
We create the part position estimator by extending the STN's architecture.
Specifically, the part position estimator utilizes a 5-layer CNN connected to a 2-layer MLP as the localization network.
The part position estimator takes a batch of fake layers $I_{B,L \times C,H,W}$ as input and estimates the position of part in each layer $\theta_{B,L \times 2}$.
Although it is possible to learn an arbitrary affine transformation, we only consider each layer's translation transformation in this work.
From the part position estimator output, we can define the affine transformation for each image layer as follows, where $i$ represents an image layer.
\begin{equation}
    \textbf{A}_{\theta}^i = \begin{bmatrix}1 & 0 & \theta_{1}^i \\0 & 1 & \theta_{2}^i\end{bmatrix}, \textbf{A}_{\theta} = \left \{ \textbf{A}_{\theta}^1, \dots, \textbf{A}_{\theta}^L \right \}
\end{equation}
Like the local GANs, we utilize the same architecture as \red{StyleGAN2-ada~\cite{karras2020training}} for the augmentation pipeline and global discriminator.
The global discriminator learns to predict based on the pretrained renderer's outputs.
Simultaneously, the renderer is also retrained with real and fake layers during global GAN training.
Same with StyleGAN2-ada, we use \textit{non-saturating GAN loss} \cite{goodfellow2014generative} with \textit{R1 regularization} \cite{mescheder2018training} for our global GAN as well.
To prevent the part position estimator from placing the part outside the layer's boundary, we also add \green{a \textit{L2 norm}} penalty if the position $\theta$ exceeds the range of $[-1,1]$.
\section{Experiments}

\subsection{Dataset}
Although several Anime face datasets are available, such as \textit{Danbooru2020} \cite{danbooru2020}, \textit{Tagged Anime Illustrations} \cite{taggedanime}, and \textit{Safebooru} \cite{safebooru}, getting a multi-layer image dataset is not trivial.
Most graphic designers would not publish the PSD files so that others cannot easily copy or modify their designs, making it extremely difficult to collect a multi-layer image dataset.
We overcame the mentioned problem by utilizing the collection of \textit{Live2D models} \cite{eikanya} on GitHub.
Live2D is a software technology that generates 2D animations without frame-by-frame animation or 3D models, only by morphing the part in the image layers.
The Live2D models mentioned above are mostly taken from various game resources that utilizing Live2D.
These models became the starting point for our dataset collection, as the model itself contains many image layers.

\begin{figure}[t]
    \centering
    \includegraphics[width=\linewidth]{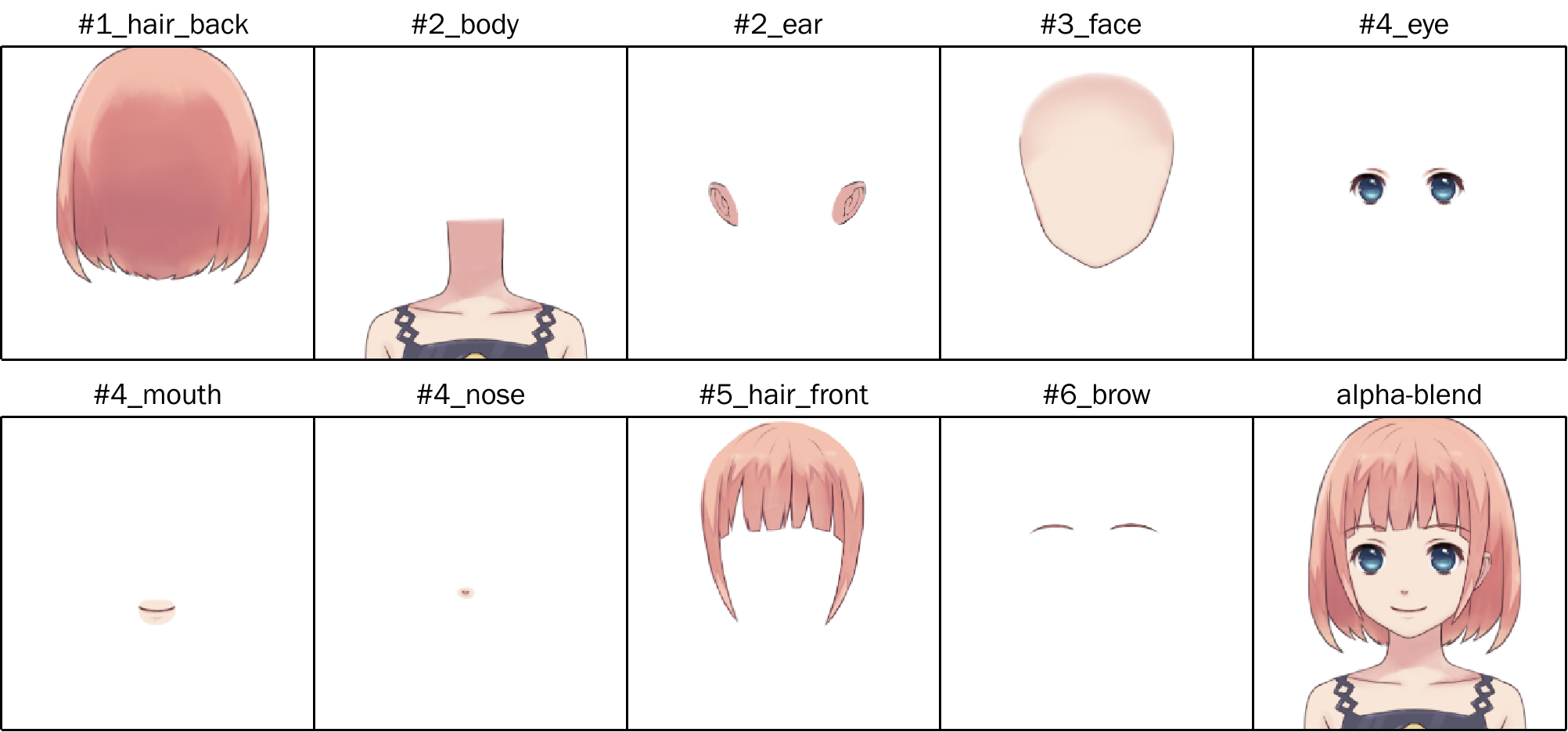}
    \caption{A sample from our dataset. The prefix number indicates the bottom-to-top order of the image layers when performing alpha blending. The image of this character is retrieved from Live2D's sample \cite{live2dsample}.}
    \label{fig:dataset_sample}
\end{figure}

We made a tool\footnote{Available at \url{https://github.com/jeffshee/live2d}} that renders a Live2D model and outputs the image layers as multiple PNG files, which allows us to train our GANs.
Since we only want to generate Anime characters' faces, we first detect and crop the face area from the image layers using \textit{lbpcascade\_animeface} \cite{lbpcascade}.
Then, we discard all the image layers that are not related to the detected face area. We resized the image layers to the resolution of \(256\times256\) pixels, upscaling with \textit{waifu2x} \cite{waifu2x} when necessary.
In most cases, the number of layers exceeds the number considered in this paper.
Therefore, we \green{merge the obtained images into nine layers} manually according to these groups: \#1\_hair\_back, \#2\_body, \#2\_ear, \#3\_face, \#4\_eye, \#4\_mouth, \#4\_nose, \#5\_hair\_front, and \#6\_brow.
Through these processes, we managed to collect 1,022 samples of multi-layer images.
However, the dataset is currently lacking in variety, as training samples are primarily originated from a game called \textit{BanG Dream! Girls Band Party!}.
Fig.\ref{fig:dataset_sample} shows a sample from our dataset.

\subsection{Experiment details}
We used a learning rate of 0.0025, a batch size of 32 (through gradient accumulation of 8 rounds, each round with batch size of 4), and \textit{Adam optimizers} \cite{kingma2014adam} for local GANs and global GAN.
The parameter of R1 regularization, $\gamma$ was set to 10.
During the step 1, the local GANs were pretrained for 2500 \textit{kimg} (thousand of real images seen by discriminator).
As we utilize nine sets of local GAN in our model, the required GPU memory is massive.
To overcome this issue, we cut down the memory usage by using different convolution configurations for each local GANs.
Specifically, we used a smaller convolution for small image layers, for example, the image layer of eyes, nose, and mouth.
Table \ref{tab:conv_config} shows the convolution configurations for each synthesis network in local generators.
We used two convolution layers for each 2x up-sampling.
The convolution configurations for each local discriminator are the likewise as its counterpart, except it performs 2x down-sampling instead.
That is, the initial and target resolutions listed in Table \ref{tab:conv_config} are reversed for the discriminators. 

\begin{table}[t]
    \centering
    \caption{Convolution configurations for each synthesis network in local generators.}
    \begin{tabular}{||l c c c||}
    \hline
    Layer & \thead{Initial\\resolution\\(pixels)} & \thead{Target\\resolution\\(pixels)} & \thead{Number of\\2x up-sampling}\\
    \hline\hline
    \#1\_hair\_back & \(8\times8\) & \(256\times256\) & 5 times \\ 
    \hline
    \#2\_body & \(8\times8\) & \(256\times256\) & 5 times \\ 
    \hline
    \#2\_ear & \(10\times14\) & \(160\times224\) & 4 times \\ 
    \hline
    \#3\_face & \(8\times8\) & \(256\times256\) & 5 times \\ 
    \hline
    \#4\_eye & \(6\times10\) & \(96\times160\) & 4 times \\ 
    \hline
    \#4\_mouth & \(6\times6\) & \(64\times96\) & 4 times \\ 
    \hline
    \#4\_nose & \(8\times4\) & \(64\times32\) & 3 times \\ 
    \hline
    \#5\_hair\_front & \(8\times8\) & \(256\times256\) & 5 times \\ 
    \hline
    \#6\_brow & \(4\times10\) & \(64\times160\) & 4 times \\ 
    \hline
    \end{tabular}
    \label{tab:conv_config}
\end{table}

\subsection{Qualitative analysis}
\begin{figure*}
    \centering
    \begin{subfigure}[b]{\textwidth}
        \centering
        \includegraphics[width=\textwidth]{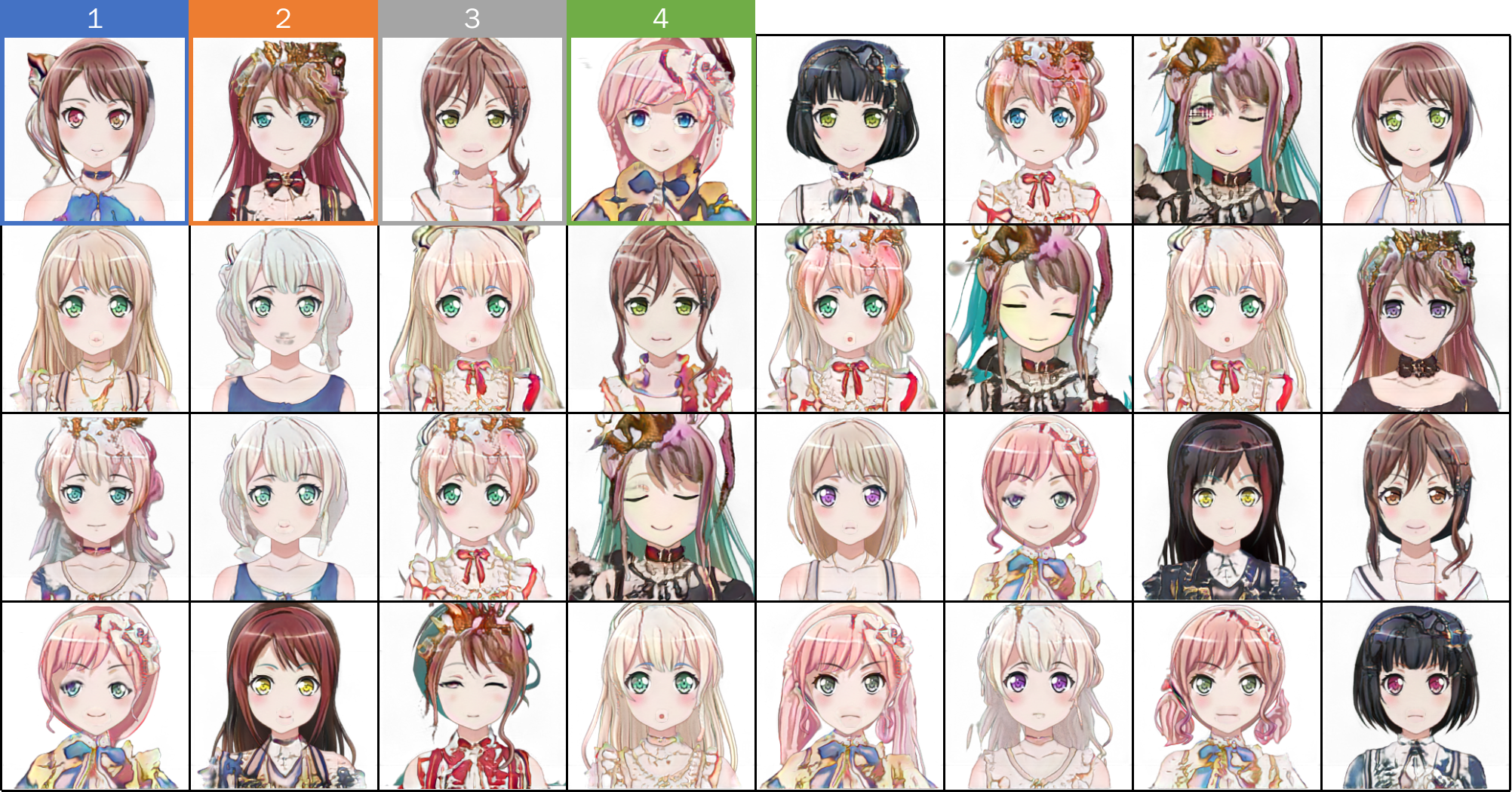}
        \caption{The alpha-blended generated samples. The \red{annotations 1$\sim$4} show the pickup samples.}
        \label{fig:fake}
     \end{subfigure}
     \hfill
     \begin{subfigure}[b]{\textwidth}
        \centering
        \includegraphics[width=\textwidth]{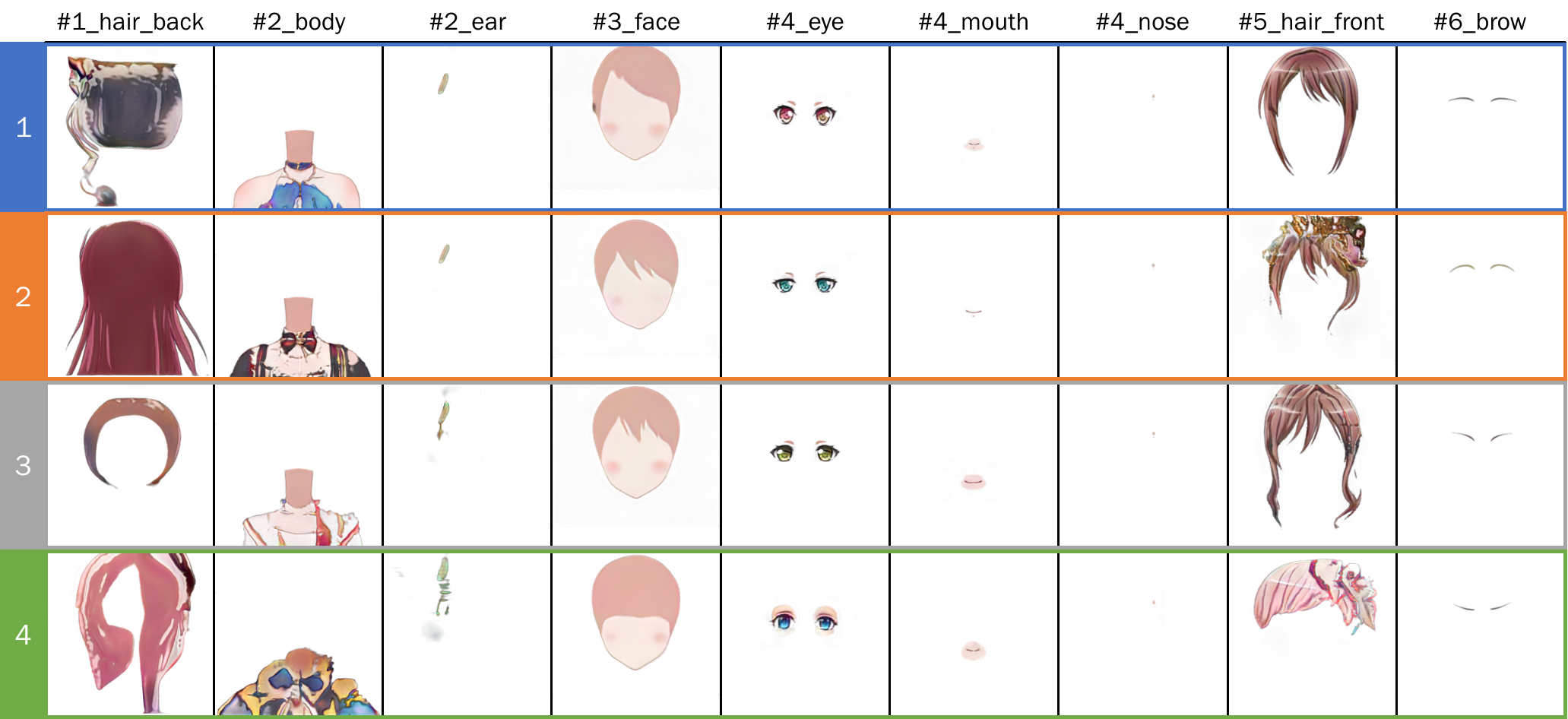}
        \caption{The layer-by-layer views of the pickup samples from Fig. \ref{fig:fake}. 
\red{Each sample is a multi-layer image that consists of nine layers.}}
        \label{fig:fake_layer}
     \end{subfigure}
     \caption{The generated samples of MontageGAN. The result is from 2600 kimg. (At step 1, local GANs were pretrained for 2500 kimg. At step 2, the global GAN was trained for 100 kimg.)  \red{The position of the part in each layer is generally well-estimated. Also, the generated image layers are generally coherent with each other.}}
     \label{fig:generated}
\end{figure*}

Fig. \ref{fig:generated} shows the generated samples of MontageGAN.
The samples shown in Fig. \ref{fig:fake} are all multi-layer images.
As a prove, Fig. \ref{fig:fake_layer} shows the layer-by-layer views of the pickup samples.
Despite the small dataset, MontageGAN can already produce some reasonably-looking faces.
The generated image layers are generally coherent with each other, indicating the global GAN had learned the global styles successfully.
For example, from Fig. \ref{fig:fake_layer}, we can find that the color and style of \#1\_hair\_back and \#5\_hair\_front seem to match each other, which did not happen during the pretraining of local GANs.
We also find that the position of part in each layer is generally well-estimated, except for the layer such as \#2\_ear and \#4\_nose.
The part in layer \#2\_ear is relatively hidden compared to other layers, as it often exists underneath the layers such as \#5\_hair\_front.
Also, the part in layer \#4\_nose is relatively small compared to other layers.
We hypothesize that layers with small or hidden parts might be relatively challenging to learn due to less noticeable appearance changes.

\subsection{Quantitative analysis}
For quantitative analysis, we compare the \textit{Fréchet Inception Distance} (FID) \cite{heusel2017gans} and \textit{Inception Score} (IS) \cite{salimans2016improved} of MontageGAN and StyleGAN2-ada.
Since both metrics use a pretrained \textit{Inception-v3} \cite{szegedy2016rethinking} model, they only support the evaluation of single-layer RGB images.
To calculate both metrics in our experiments, we first alpha-blended both real and fake layers into single-layer images, then added a white background to all images, making them RGB images.
Since there is no other comparison method for generating multi-layer images, we extended the implementation of StyleGAN2-ada (\textit{modified StyleGAN2-ada}), allowing it to generate multi-layer images.
Specifically, the generator of modified StyleGAN2-ada generates a tensor $I_{B,L\times C,H,W}$ as layers.
Correspondingly, the discriminator of modified StyleGAN2-ada also predicts based on the input tensor $I_{B,L\times C,H,W}$.
Based on the single-layer image performance, Table \ref{tab:comparison} shows FID and IS's comparison results against our dataset.

For IS, it shows that both MontageGAN and StyleGAN2-ada give comparable results. 
On the other hand, FID shows that MontageGAN's single-layer image performance is suboptimal compared to StyleGAN2-ada.
We hypothesize that MontageGAN, which generates multi-layer images, has an inevitable rise in complexity that causes the drop of single-layer image performance.

On the plus side, MontageGAN achieved the best performance among the multi-layer image generation methods.
Through our experiment, the modified StyleGAN2-ada could not even learn stably; it had a severe mode collapse issue and the training also started to collapse at around 2100 kimg due to the discriminator overfit.
This suggests the importance and effectiveness of our renderer in limiting the excess information of multi-layer images fed to the discriminator.

\begin{table}[t]
    \centering
    \caption{Comparison using FID and IS against our dataset, based on the single-layer image performance. $\downarrow$ indicates lower is better. $\uparrow$ indicates higher is better. Red and blue fonts indicates the best and the 2\textsuperscript{nd} best performance, respectively.}
    \begin{tabular}{||l l c c||}
    \hline
    Method & \thead{Method type} & \thead{FID$\downarrow$} & \thead{IS$\uparrow$}\\
    \hline\hline
    StyleGAN2-ada\footnotemark[4] & Single-layer & \textcolor{red}{37.74} & \textcolor{red}{1.88} \\ 
    \hline
    Modified StyleGAN2-ada\footnotemark[5] & \textbf{Multi-layer} & 137.57 & 1.77 \\ 
    \hline
    MontageGAN (Our method) & \textbf{Multi-layer} & \textcolor{blue}{56.65} & \textcolor{red}{1.88} \\
    \hline
    \end{tabular}
    \label{tab:comparison}
\end{table}
\footnotetext[4]{StyleGAN2-ada's is result from 2620 kimg, which is the same extent to MontageGAN's total training duration.}
\footnotetext[5]{Modified StyleGAN2-ada's result is from 2000 kimg, which is about 100 kimg before the training collapsed.}
\section{Conclusion}
In this work, we proposed MontageGAN, a GANs framework specializing in multi-layer image generation.
We applied MontageGAN to Anime face image generation and showed its potential of generating multi-layer images.
We also showed our method to collect a multi-layer Anime face dataset that makes the training of MontageGAN possible.

Our future work would be fine-tuning our model to improve the image quality.
One idea is to incorporate \textit{gated convolution} into the synthesis networks of local generators.
As the image layers will be alpha-blended to form the finished image, there are many transparent pixels in each layer image.
When generating such a layer image, we believe it is more efficient to consider the image's alpha channel as a soft-gate instead of a color value like RGB, which adds another dimension of complexity to the training process.

\bibliographystyle{IEEEtran}
\bibliography{myref}

\section*{Appendix}

\setcounter{figure}{0}
\setcounter{table}{0}
\setcounter{section}{0}
\renewcommand{\thesection}{\!A\arabic{section}}
\renewcommand{\thefigure}{\!A\arabic{figure}}
\renewcommand{\thetable}{\!A\arabic{table}}
\renewcommand{\topfraction}{.9}
\renewcommand{\bottomfraction}{.9}
\renewcommand{\textfraction}{.01}
\renewcommand{\floatpagefraction}{.9}

\section{Layer-wise post-editing}
\begin{figure*}[!t]
    \centering
    \includegraphics[width=\linewidth]{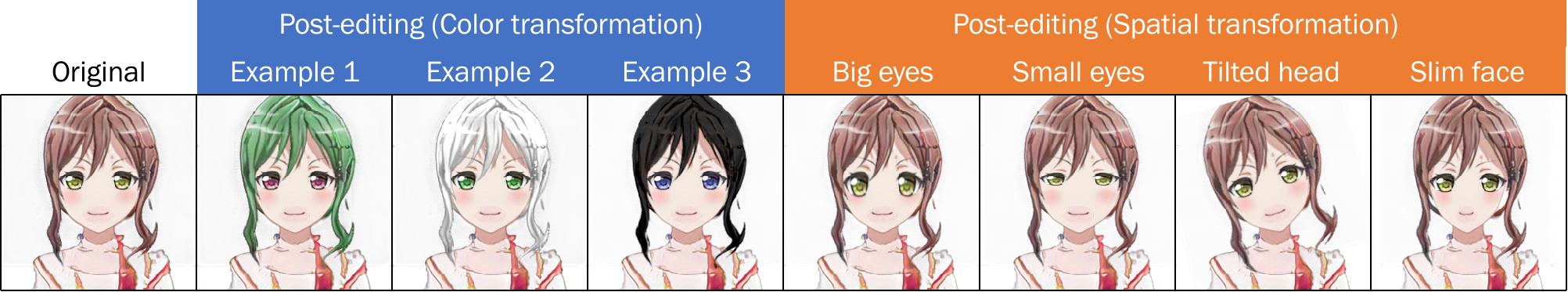}
    \caption{Examples of manual post-editing by applying color and spatial transformations to specific image layers of global GAN's output. The leftmost is the original, taken from Fig.~\ref{fig:generated}. We utilized GIMP for the post-editing.}
    \label{fig:post-edit}
\bigskip
    \centering
    \includegraphics[width=\linewidth]{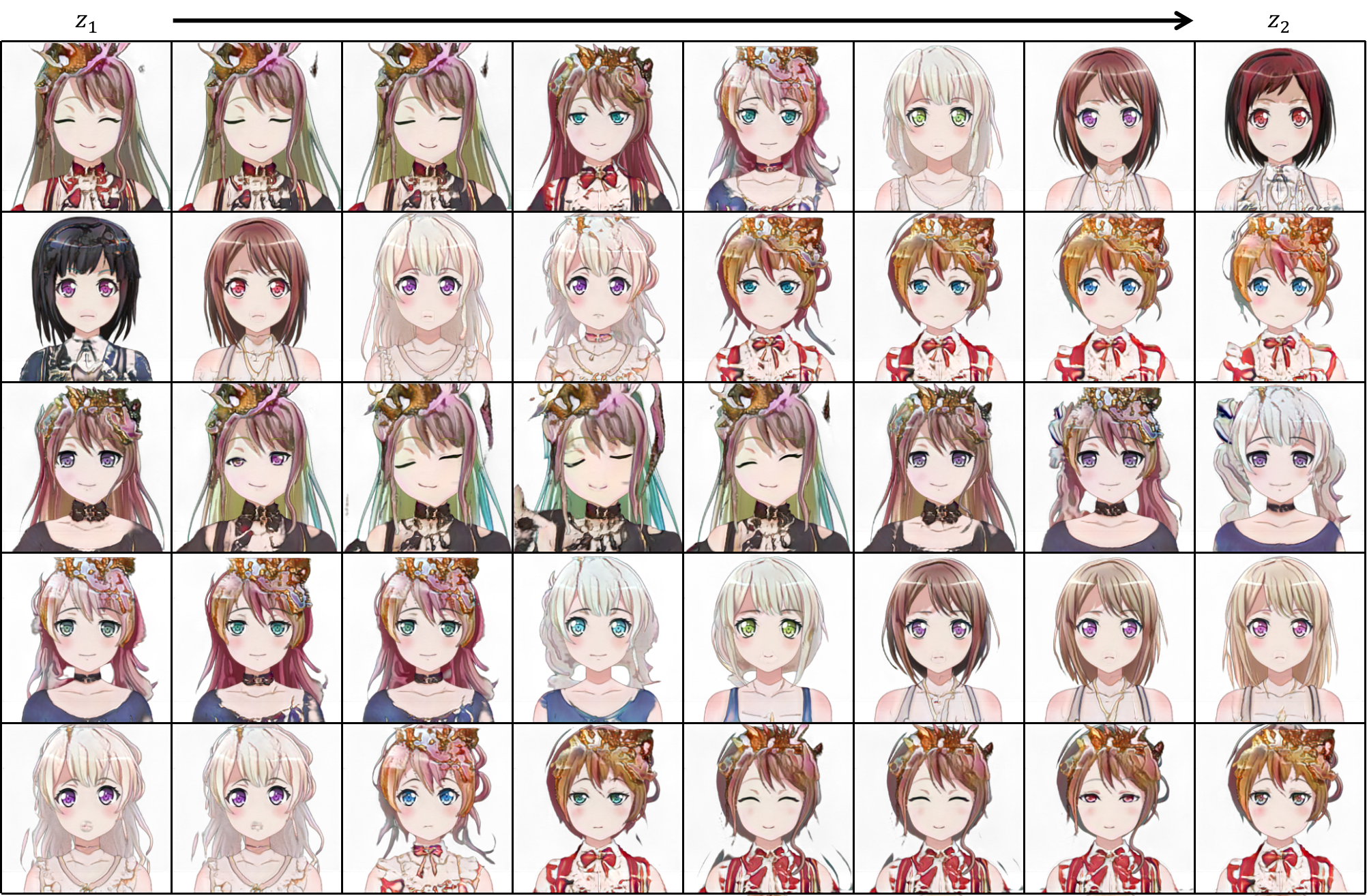}
    \caption{Latent space interpolation between 5 pairs of latent variables, denoted as $z_1$ and $z_2$. The rows represent different instances of latent variables pairs, and the columns represent the image layers generated with the interpolated latent variables.
    The smooth interpolation of latent space shown in the result indicates that MontageGAN had successfully learned a meaningful representation of the layer images.
    }
    \label{fig:interpolation}
    \vspace{1cm}
\end{figure*}

The key feature of MontageGAN over traditional GANs is the multi-layer images generation that makes \textit{layer-wise post-editing} a viable task.
{Fig.~\ref{fig:post-edit}} shows examples of manual post-editing by applying color and spatial transformations to some image layers of the global GAN's output. Thanks to the nature of multi-layer images, we can easily apply color and spatial transformations to specific image layers to change the corresponding parts' color and geometry, respectively.

\begin{table*}[t!]
    \centering
    \caption{Comparison of functions.}
    \begin{tabular}{|l l ll l ||c c|}
    \hline
     & &\multicolumn{2}{l}{\thead{Post-editing}} & \thead{Contextual} &  & \\ \cline{3-4}
        Method & \thead{Method type}&
        \thead{Geometric} & \thead{Color}& \thead{analysis} & \thead{FID$\downarrow$} & \thead{IS$\uparrow$}\\
    \hline\hline
    StyleGAN2-ada & Single-layer & Unable & Unable & Unable & \textcolor{red}{37.74} & \textcolor{red}{1.88} \\ 
    \hline
    Modified StyleGAN2-ada & \textbf{\textcolor{red}{Multi-layer}} & \textcolor{red}{Able} &\textcolor{red}{Able} &\textcolor{red}{Able}  &137.57 & 1.77 \\ 
    \hline
    MontageGAN (Our method) & \textbf{\textcolor{red}{Multi-layer}}   & \textcolor{red}{Able} & \textcolor{red}{Able} & \textcolor{red}{Able} & \textcolor{blue}{56.65} & \textcolor{red}{1.88} \\
    \hline
    \end{tabular}
    \label{tab:functional-comparison}
\end{table*}

\section{Generation by interpolation}
Like other GAN-based image generation methods, MontageGAN allows ``generation by interpolation.''
{Fig.~\ref{fig:interpolation}} shows the latent space interpolation between 5 pairs of latent variables.
We calculate the linear interpolation for each pair of latent variables $z_1$ and $z_2$, then generate the image layers with the interpolated latent variables.
This figure shows the case where individual layers are interpolated; if necessary, we can interpolate facial parts at certain layers by utilizing our multi-layer generation framework. More specifically, by fixing some layers and interpolating other layers, we can make, for example, an eye-blinking sequence of images while keeping the facial parts other than eyes.

\section{More about our comparative study}
In Table \ref{tab:comparison}, StyleGAN2-ada achieved the best FID among the three methods; however, as we emphasized in Table~\ref{tab:functional-comparison}, there are clear functional differences between StyleGAN2-ada and ours.
StyleGAN2-ada is not a multi-layer image generation method.
In contrast, as shown by {Fig.~\ref{fig:post-edit}}, ours is multi-layer and therefore it is possible to do post-editing and contextual analysis after generation. 
\par
The reason why we provide StyleGAN2-ada's FID is to show how close a multi-layer image generation method is to a single-layer method, StyleGAN2-ada, in the quality of the alpha-blended appearance of generated samples. (The result in Table~\ref{tab:comparison} shows that they achieved the same IS value, 1.88.)
Note that MontageGAN achieved better FID and IS than the other multi-layer image generation method, Modified StyleGAN2-ada. Also note that there is no well-known SOTA method for multi-layer image generation as the time of writing -- this is because, to the authors' best knowledge, our MontageGAN is a pioneering work of multi-layer image generation.

\section{The number of channels}
Since the training and generated samples consist of RGBA channels, the number of channels used as discriminator's inputs is also 4.
Inside the GANs, we set the number of CNN's channels according to their input resolution (the longest side for a rectangle). 
The higher the input resolution, the fewer channels we used to conserve the calculation resource.
Specifically, we used 512, 256, 128, and 64 channels for input resolutions of 4$\sim$32 pixels, 33$\sim$64 pixels, 65$\sim$128 pixels, and 129$\sim$256 pixels, respectively.

\end{document}